\documentclass[a4paper]{article}
\usepackage[utf8]{inputenc}
\usepackage[T1]{fontenc}
\usepackage{fixltx2e}
\usepackage{graphicx}
\usepackage{longtable}
\usepackage{float}
\usepackage{wrapfig}
\usepackage{soul}
\usepackage{textcomp}
\usepackage{marvosym}
\usepackage{wasysym}
\usepackage{latexsym}
\usepackage{amssymb}
\usepackage{hyperref}
\title{On measuring linguistic intelligence}
\author{Maxim Litvak (maxim@litvak.eu)}
\date{\today}
\hypersetup{
  pdfkeywords={},
  pdfsubject={},
  pdfcreator={Emacs Org-mode version 7.9.3f}}

\begin{document}

\maketitle

\begin{center}
Abstract
\end{center}

This work addresses the problem of measuring how many languages a person ``effectively'' speaks given that some of the languages are close to each other. In other words, to assign a meaningful number to her language portfolio.

Intuition says that someone who speaks fluently Spanish and Portuguese
is linguistically less proficient compared to someone who speaks fluently
Spanish and Chinese since it takes more effort for a native Spanish speaker
to learn Chinese than Portuguese. 
As the number of languages grows and their proficiency levels vary, it gets even more complicated to assign a score to a language portfolio.

In this article we propose such a measure (”linguistic quotient” - LQ) that can account for these effects.

We define the properties that such a measure should have. They are based on the idea of coherent risk measures from the mathematical finance.

Having laid down the foundation, we propose one such a measure together with the algorithm that works on languages classification tree as input.

The algorithm together with the input is available online at \hyperref[lingvometer.com]{lingvometer.com}
\newpage
\tableofcontents
\section{Introduction}
\label{sec-1}

If we aim to compare different language portfolios (consisting of
different languages at different proficiency levels), we need some way to
measure it. Similar to IQ that is supposed to measure the
intelligence, we need a ``LQ'' that measures linguistic intelligence.

Linguistic intelligence for the scope of this article is the achieved
proficiency in some set of languages, i.e. not the potential to learn fast
a new one, but an achievement that already took place.  

The main idea is that languages that are related to each other give less to the linguistic intelligence than those that are unrelated.  

We will deploy actively the ideas and methods from finance. A sample of languages resembles to some extent a portfolio of assets. 

To a given portfolio of assets it is more valuable to add an asset that is not correlated or even negatively correlated to the assets of a given portfolio (e.g. [Markowitz 52]). 

In the same way for a given sample of languages (further, portfolio of
languages) it is more valuable (in the sense of linguistic intelligence) to add a language
that is not related to the languages of a given portfolio.
  
The article is organized as follows. In the section \hyperref[sec-first]{3.1}  we design the measure
given the set of properties that sounds reasonable from the intuitive and logical point of view.
  
In the section \hyperref[sec-lq]{3.2} we turn this design into mathematical model that allows to calculate a score to a given portfolio of languages (it could be calculated for any language profile at \hyperref[lingvometer.com]{lingvometer.com}). 
\section{Properties of linguistic intelligence}
\label{sec-2}

We reason on properties the linguistic intelligence should
have.  Based on these principles we will later derive the formal rules.
  
We consider all the languages as equal to each other. 
It is not a trivial assumption since there are languages that have been developing over
thousands of years a reach literary and oral traditions, serving as medium
for the scientific expression etc.  

There are other languages that fail to compete with the former on this
account.  It is tempting to say that they are less valuable. However, we still regard them as equal since
the ”undeveloped” languages could be harder to acquire exactly because
they don’t have the written tradition.  

People master the languages to the different extent. Assume that
the proficiency degree of the grown-up educated native speaker
is 1. Someone who never faced it has the level 0 in this language. It
would be logical to demand that increasing the proficiency in some
language of the portfolio would increase the LQ of the person.

We set the measure of the portfolio of \emph{n} independent languages to be \emph{n},
i.e. LQ of the portfolio of Spanish and Chinese would be 2.

Consequently, measure of the portfolio of \emph{n} languages, such that
some of them are related to each other is less than \emph{n}.  Further, if
we add to a portfolio a language that is already there then the measure of
the portfolio must remain the same.

Thus, we would assign to the portfolio of two languages a number between 1 and 2. The closer they are, the closer is this number to 1.
The further they are, the closer is this number to 2. 

We interpret such a measure as the ``real'' (or effective) number of languages a person speaks. 

Thus, one and one is two if we add Spanish to Chinese. If we add
Spanish and Portuguese, then one and one could be something like 1.3, maybe
a little bit less or a little bit more.

There are also other reasons than the time to learn a new language to assign a higher score to Spanish+Chinese than to Spanish+Portuguese. We could also argue that learning a distant language, one learns also new structures that are not there in the related language.

\newpage
\section{Formal approach}
\label{sec-3}

In this chapter we will argue on reasonable properties of a linguistic intelligence measure. Further, we formalize the properties as axioms and propose a formula that satisfies them.
\subsection{First principles}
\label{sec-3-1}
\label{sec:first}

We write our considerations on how the LQ measure should behave in the following
list of axioms. It was inspired by the idea of coherent risk measures [Artzner et al. 99] from financial mathematics.

Consider a language \emph{l} is an element of languages space \emph{L} and weighted language \emph{w} (in other words a language with proficiency level) is an element of the space \emph{W}:=$L\times[0,1]$ that is a language and a number between 0 and 1 (1 is a fluent command of a language, 0 means any knowledge is absent).

Portfolio $\Pi$ of N languages is an element of space $W^N$. If needed we could also consider the portfolio $\Pi$ of N languages to be an element of space $L^N$ (i.e. languages at fluent proficiency).

\textbf{Definition}: A \emph{linguistic intelligence measure} (l.i.m.) is a function s.t. $\lambda:W^N\rightarrow R$

If we set all proficiency levels to 1, then l.i.m. is also defined on $L^N$: $\lambda: L^N\rightarrow R$.

Now, we formalize the arguments of the previous chapter as axioms.

We consider all languages to be equal:    

\textbf{Axiom E}. Equivalence. $\forall l\in L, l\neq\emptyset: \lambda(l)=1$

A portfolio of languages weights at most as sum of its components:    
   
\textbf{Axiom S}. Subadditivity. For any 2 language portfolios it must hold:    
    
$\forall \Pi_1,\Pi_2 \in W^N:\ \lambda(\Pi_1\cup\Pi_2 )\leq \lambda(\Pi_1)+\lambda(\Pi_2)$

\textbf{Axiom ND}. No double-counting. For a language \emph{l} that is in portfolio $\Pi$ it must
hold:    

$l\in\Pi \Rightarrow \lambda(\Pi\cup l)=\lambda(\Pi)$

\textbf{Axiom I}. Independency. For a language \emph{l} that is independent to any language
in portfolio $\Pi$ it must hold:    

$\Pi\perp l \Rightarrow \lambda(\Pi\cup l)=\lambda(\Pi)+\lambda(l)$

\textbf{Axiom PH}. Positive homogeneity. Proficiency effect is linear.    

$\forall c\in [0,1], \forall l\in L,\ w=(c,l)\in W: \lambda(w)=c\lambda(l)$

\textbf{Definition} A linguistic measure is called \emph{coherent} (c.l.i.m.) if it satisfies axioms E, S, D, I, PH.

There are many measures that satisfy these axioms. We propose one in the next chapter.

Even though a c.l.i.m. seems to be strictly defined since there are many requirement to be met, there are 2 key points in the whole construction that still leave the room for interpretation: how the language space is constructed and how the set-theoretic operations (independence and union) between elements of this space are defined. Later on, we will see 2 completely different construction approaches and further constructions are still possible.

Further, we observe the axioms from different points of view to understand them better. If the axiom E takes place, then axioms S, D and I lead to the following inequality:
\[
\forall \Pi \in L^N, \forall l \in L: \lambda(\Pi) \leq \lambda(\Pi\cup l)  \leq \lambda(\Pi)+1 
\]
Thus, adding a language to a portfolio adds a number between 0 and 1 to linguistic intelligence. 

This inequality helps us to reduce the range of l.i.m., s.t. $\lambda: L^N\rightarrow [1,N]$ or in terms of W (with PH axiom) $\lambda: W^N\rightarrow [0,N]$.

Simplifying further the inequality, we can write it for a portfolio of 2 languages as follows:
\[
\forall l_1,l_2 \in L, \Phi=(l_1\cup l_2)\in L^2: 1 \leq \lambda(\Phi) \leq  2
\]

If for some reason a space L is defined in such a way that it includes an empty element, then the axiom E must be adjusted such that $\lambda(\emptyset)=0$
\subsection{LQ}
\label{sec-3-2}
\label{sec:lq}

We consider the portfolios of languages to be a tree with hypothetical Tower of Babel (ToB) language as a source. A portfolio consisting of only one language would be a path from the source to the language through the language families, groups, subgroups etc classification.

The children of ToB are language families like Indo-European or Sino-Tibetan. This is the layer of nodes of rank 1, the languages are considered independent if they belong to different language families.

Thus, the language space $L^N$ is the full tree of all (N) languages. The languages are the leaves of the tree and portfolios are the induced subgraphs of the full graphs containing the source (ToB), leaves and the paths between them. Thus, a portfolio of 1 language would be a path from the source through all families, sub-families, groups etc to this language. The union of 2 portfolios would be the union of the corresponding subgraphs.

To illustrate it, consider a portfolio $\Pi$ consisting of Chinese and Serbian and portfolio $\Phi$ consisting of English and Slovene. Then the unified portfolio $\Psi=\Pi\cup\Phi$ would contain all 4 languages. The situation is illustrated on the figure \hyperref[fig-union]{1}.

\begin{figure}[htb]
\centering
\includegraphics[scale=0.3]{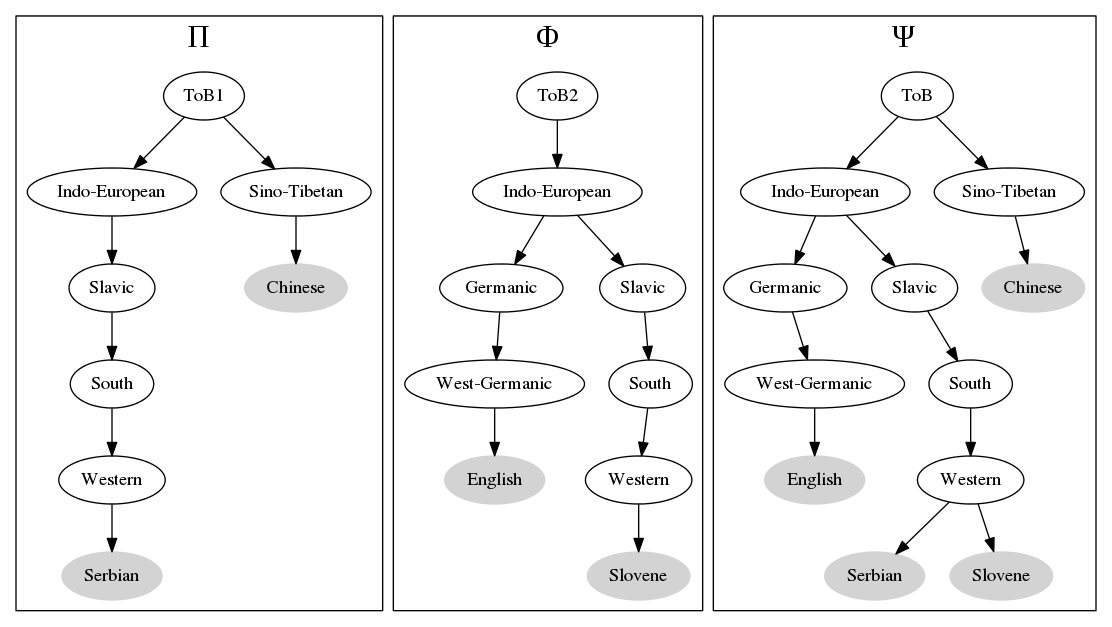}
\caption{\label{fig:union}Union of portfolios}
\end{figure}

Let N be the maximal depth of the tree and $V_r$ the set of all nodes of depth \emph{r}. Initialize all languages with their proficiencies (not necessarily all hanging in the deepest layer).

Starting from the deepest layer up to the source calculate for each node in the layer iteratively the LQ (bottom-up). 

Denote the LQ of the node \emph{v} as $\lambda_v$ and Ch(v) as the set of nodes which are the children of node v.
For $r=\overline{N,0}$,  $\forall v \in V_r$

\begin{center}
\[
        \lambda_v=\big(\sum_{c\in Ch(v)} \lambda_c^{\sqrt{r}} \big)^{\sqrt{\frac{1}{r}}}
\]
\end{center}

We define the result of the last step of this iterative process as LQ measure. The pseudocode for the algorithm both the straightforward way and the recursive way is given in the appendix.

\textbf{Definition}. \emph{LQ}-measure is such a l.i.m. that $LQ=\lambda_v, v\in V_0$

There's only one node at the layer of depth 0 that is the root of the tree (Tower of Babel language). A numerical example is given in Appendix.

In the formula, we take the LQs of the nodes to the power of the root of the rank and not the root itself as on the real data it produces more intuitive results. Actually, taking instead of the root of the rank any monotone function of the rank with the fixed point at 1 (i.e. $f(1)=1$) would give another c.l.i.m.:

\begin{center}
\[
        \lambda_v=\big(\sum_{c\in Ch(v)} \lambda_c^{f(r)} \big)^{\frac{1}{f(r)}}
\]
\end{center}

LQ turns out to be coherent.

\textbf{Lemma}. LQ is c.l.i.m.

\textbf{Proof}. Axiom S. Consider the language portfolios $\Phi$ and $\Pi$ that could have common elements (they have at least ToB in common). We could rewrite the children of the node as being the union of its children belonging to $\Pi$ and its children belonging to $\Phi$, then we algorithm to calculate $LQ(\Pi\cup\Phi)$ looks as follows (again starting with the nodes from the bottom):

\[
        \lambda_v=\big(\sum_{c\in Ch^\Pi(v)\cup Ch^\Phi(v)} \lambda_c^{f(r)} \big)^{\frac{1}{f(r)}} 
\]

The algorithms to calculate $LQ(\Pi)$ and $LQ(\Phi)$ are respectively

\[
        \lambda_v=\big(\sum_{c\in Ch^\Pi(v)} \lambda_c^{f(r)} \big)^{\frac{1}{f(r)}} 
\]

and

\[
        \lambda_v=\big(\sum_{c\in Ch^\Phi(v)} \lambda_c^{f(r)} \big)^{\frac{1}{f(r)}} 
\]

The numbers $\lambda(c)$ forms a vector from the space $R^{\#Ch(v)}$, their Minkowski distance to the 0 vector with $p=f(r)$ is $\lambda_v$. In case $Ch^\Phi(v)\cap Ch^\Pi(v)=\emptyset$ the triangular inequality states that

\[
        \lambda_v=\big(\sum_{c\in Ch^\Pi(v)\cup Ch^\Phi(v)} \lambda_c^{f(r)} \big)^{\frac{1}{f(r)}} \leq \big(\sum_{c\in Ch^\Pi(v)} \lambda_c^{f(r)} \big)^{\frac{1}{f(r)}} +\big(\sum_{c\in Ch^\Phi(v)} \lambda_c^{f(r)} \big)^{\frac{1}{f(r)}}
\]

If the intersection of the children for this node is not empty, then we have additional terms on the right side of the inequality and, thus, still holds. 
Note that on the right side of inequality we have the components that flow in calculation of $LQ(\Pi)$ and $LQ(\Phi)$ respectively.

Thus, going from the bottom to the top, we will finally have

\[
LQ(\Pi\cup\Phi) \leq LQ(\Pi) + LQ(\Phi)
\]

This satisfies the axiom S.

Axiom I is satisfied due to the property of the formula that at the rank 1 (this level is considered to contain independent entities) the sum of LQs is linear (i.e. at no cost).

Axioms E and PH are satisfied due to the fact that initialized with 1 (or with $c\in[0,1]$ in case PH) the LQ of the only language could be push up alone the path to the source at no cost. 

Axiom ND is satisfied due to the space construction.
$\blacksquare$

A shortcoming of the approach is that some language dependences are not captured by tree structure. For instance, the direct French influence on English could be represented by the edge between them. This would, however, destroy the tree structure. 

One of the advantages is that the language tree data is easily available, for instance in [ethnologue 2015].

One could try the algorithm on different input at \hyperref[lingvometer.com]{lingvometer.com}
\subsection{Matrix approach}
\label{sec-3-3}

We have seen in the previous chapter the measure constructed on language trees. However, it does not have to be trees. In this chapter we will discuss an alternative approach, namely the one based on correlation matrices.

The entries of the matrices represent a correlation or language distance (more precisely 1 minus distance to make it look like a correlation) expressed as a number between 0 and 1. There are many works on measuring such a distance, to name a few [Petroni and Serva 2010], [Delsing and Åkesson 2005], [Koehn 2005], [Chiswick and Miller 2005], [Gingsburgh and Weber 2011]. They use different approaches: difficulty of acquiring a language, difficulty of machine translation between languages, number of words in common etc. Most of them cover rather a small part of all possible correlations/distances. With N languages, there must be N(N-1)/2 distances. 

The same as stocks prices are correlated with the information stored in correlation matrices, we could also consider the languages to be correlated.

We try to construct a l.i.m. on $2\times 2$ matrices i.e. portfolios of 2 languages. The correlation matrix in this case looks like this:

\begin{center}
\begin{tabular}{|cc|}
      1  &  $\rho$  \\
 $\rho$  &       1  \\
\end{tabular}
\end{center}

Denote it as M($\rho$).

Languages are independent if $\rho=0$ and more dependent as $\rho$ is closer to 1 with two languages being equal if $\rho=1$.

In order for the l.i.m. to be coherent, we need among other things $\lambda(M(0))=2$ and $\lambda(M(1))=1$ with $\lambda$ being monotone decreasing on $\rho$.

One such a c.l.i.m. is $\lambda(M(\rho))=2-\rho$. Axioms I and ND we checked above. The matrix of one language is simply 1, thus, axiom E holds as well. Axiom PH is not relevant on $L^2$ space (it's relevant only on $W^N$ spaces). Axiom S holds since $2-\rho\leq 1 + 1$ ($0\leq\rho\leq 1$).

It is not the only c.l.i.m. on this space since a familty of c.l.i.m. could be constructed like this $\lambda(M(\rho))=2-\rho^r, r>0$

We discussed a particular case that shows an example of c.l.i.m. on the matrix space.
The general case is still open.

One of the shortcoming of the matrix approach is that the data on all $N\times (N-1)$ dependencies is not available. Different studies assign if close, but still different numbers. Apparently, the tree data could be mapped to the matrix and vice versa (e.g. [Petroni and Serva 2008]).
\section{Conclusion}
\label{sec-4}

We presented a sound way to measure a portfolio of languages. How could it be used except for measuring someone's linguistic intelligence?

There is a broad field of research on intersection between economics and linguistics. An extensive overview could be found in [Grin 2003]. 

LQ could be used for instance as communication cost function. An institut that evaluates several options as a working language(s), could aim to minimize the overall LQ of its members since it would also mean the minimization of communication costs. For example, the optimal language (or an optimal bundle of languages) for European Union could be chosen in such a way that the aggregated LQ of European population would be minimal.

The implementation of the LQ algorithm could be found at \hyperref[lingvometer.com]{lingvometer.com}
\section{References}
\label{sec-5}

[Artzner et al 1999] Artzner, P., Delbaen, F., Eber, J. M., \& Heath, D. (1999). Coherent measures of risk. Mathematical finance, 9(3), 203-228.

[Markowitz 1952] Markowitz, H. (1952). Portfolio selection. The journal of finance, 7(1), 77-91.

[Grin 2003] Grin, F. (2003). Language planning and economics. Current Issues in Language Planning, 4(1), 1-66.

[ethnologue 2015] Lewis, M. Paul, Gary F. Simons, and Charles D. Fennig (eds.). 2015. Ethnologue: Languages of the World, Eighteenth edition. Dallas, Texas: SIL International. Online version: \href{http://www.ethnologue.com}{http://www.ethnologue.com}.

[Petroni and Serva 2010] Petroni, F., \& Serva, M. (2010). Measures of lexical distance between languages. Physica A: Statistical Mechanics and its Applications, 389(11), 2280-2283.

[Delsing and Åkesson 2005] Delsing, L. O., \& Åkesson, K. L. (2005). Håller språket ihop Norden. En forskningsrapport om ungdomars förståelse av danska, svenska och norska.

[Petroni and Serva 2008] Petroni, F., \& Serva, M. (2008). Language distance and tree reconstruction. Journal of Statistical Mechanics: Theory and Experiment, 2008(08), P08012 

[Koehn 2005] Koehn, P. (2005, September). Europarl: A parallel corpus for statistical machine translation. In MT summit (Vol. 5, pp. 79-86).

[Chiswick and Miller 2005] Chiswick, B. R., \& Miller, P. W. (2005). Linguistic distance: A quantitative measure of the distance between English and other languages. Journal of Multilingual and Multicultural Development, 26(1), 1-11.

[Ginsburgh and Weber 2011] Ginsburgh, V., \& Weber, S. (2011). How many languages do we need?: The economics of linguistic diversity. Princeton University Press.
\newpage
\section{Appendix}
\label{sec-6}
\subsection{Pseudocode for LQ algorithm}
\label{sec-6-1}

The pseudocode is presented in python style. First, initialize all leaves with LQ equal to the proficiency level of the corresponding language. Note that the leaves could lie in different layers and, thus, have different rank:

\begin{verbatim}
for node in nodes:
        if node is leaf:
                node.lq=node.language.proficiency
\end{verbatim}

The calculation:

\begin{verbatim}
for rank in range[deepest_rank,0]: # i.e. start with deepest_rank, end with 0
        for node in nodes_of_layer(rank):
                for child in children(node):
                        node.lq+=sqrt(child.lq)
                node.lq=power(node.lq, 1/sqrt(rank))
\end{verbatim}

The result is

\begin{verbatim}
LQ=nodes_of_layer(0)[0].lq
\end{verbatim}
                
We also write down the recursive version of the algorithm. The initialization step is the same. The recursive function that will do the job could be implemented like this

\begin{verbatim}
def lq_recursive(rank):
        if children(node) is empty:
                return node.language.proficiency

        else:
                temp_sum=0
                for child in children(node):
                        temp_sum+=lq_recursive(child)
                return power(temp_sum, node.rank+1)
\end{verbatim}

Then we can calculate LQ with the following call:

\begin{verbatim}
LQ=lq_recursive(ToB)
\end{verbatim}
\subsection{Numerical Example to LQ-Tree}
\label{sec-6-2}

We introduce here an example of a language profile that is not trivial, but also not very complex. It contain patterns that test the first principles. Someone speaks fluently Serbian, Slovene, Croatian and Chinese fluently. Besides, he has some command of English that qualifies at 50\% level.
We also do the arithmetics and show every step of the calculation.
Consider the following language portfolio. To initialize the algorithm, we assign the rank to each layer and set LQ to 1 for each language except for English where we set LQ to 0.5 (Fig. 1).
\begin{figure}[htb]
\centering
\includegraphics[scale=0.25]{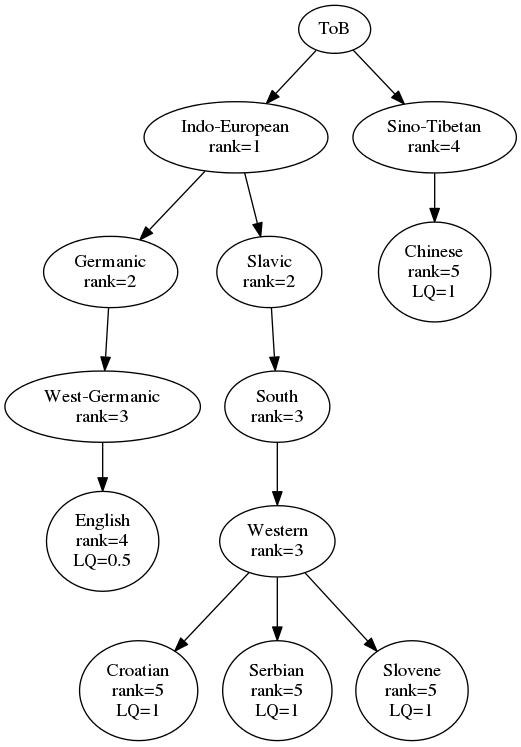}
\caption{Initialization}
\end{figure}

The deepest layer (that of Serbian, Slovene and Croatian) is of rank 5. Now, calculate the LQ for Western Branch of South-Slavic languages (Fig. 2).

\[
        \lambda_{Western}=\big( 1^{\sqrt{5}} + 1^{\sqrt{5}} +1^{\sqrt{5}} \big)^{\sqrt{\frac{1}{5}}}=3^{\sqrt{\frac{1}{5}}}\approx 1.63
\]

If a node has just one child than according to the formula it takes its LQ unbiased. 

For another example, let's take the node of the Indo-European family (rank of the layer is 2). It has 2 children, namely Germanic and Slavic groups with LQ calculated at previous iterations equal to 0.5 and 1.63 respectively. 

\[
        \lambda_{Western}=\big(0.5^{\sqrt{2}}+1.63^{\sqrt{2}} \big)^{\sqrt{\frac{1}{2}}}\approx 1.84
\]

\begin{figure}[htb]
\centering
\includegraphics[scale=0.25]{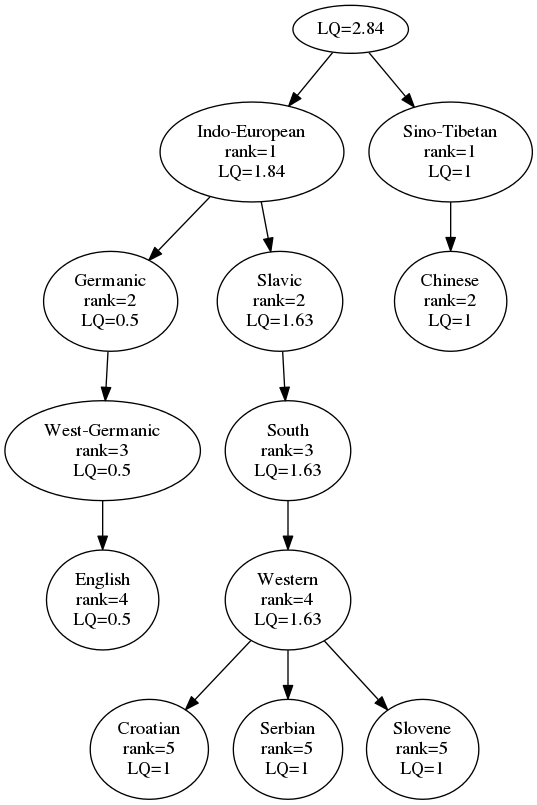}
\caption{Calculated tree}
\end{figure}

At the final step we can sum the LQs of the languages families at no cost: $LQ=1.84+1=2.84$

The whole tree with LQs for each node is represented on the \hyperref[fig-full]{Fig.3}

\end{document}